\documentclass[letterpaper, 10 pt, conference]{ieeeconf}

\IEEEoverridecommandlockouts 

\usepackage{nfpackages}
\usepackage{nfcommands}

\title{\LARGE \bf
Long-Horizon Prediction and Uncertainty Propagation with Residual Point Contact Learners}
\author{Nima Fazeli$^{1}$, Anurag Ajay$^{2}$, and Alberto Rodriguez$^{1}$
\thanks{$^{1}$Mechanical Engineering Department, MIT, Cambridge, MA, USA
        {\tt\small <nfazeli,albertor>@mit.edu}}%
\thanks{$^{2}$Computer Science and Artificial Intelligence Laboratory, MIT, Cambridge, MA, USA 
        {\tt\small aajay@mit.edu}}%
}

\begin{document}

\maketitle
\thispagestyle{empty}
\pagestyle{empty}

\begin{abstract} 
The ability to simulate and predict the outcome of contacts is paramount to the successful execution of many robotic tasks. Simulators are powerful tools for the design of robots and their behaviors, yet the discrepancy between their predictions and observed data limit their usability. In this paper, we propose a self-supervised approach to learning residual models for rigid-body simulators that exploits corrections of contact models to refine predictive performance and propagate uncertainty. We empirically evaluate the framework by predicting the outcomes of planar dice rolls and compare it's performance to state-of-the-art techniques.
\end{abstract}

\section{Introduction}

Simulators are used extensively in robotics for tasks such as planning, control, state estimation, system identification, and design. Simulators (ODE \cite{smith2005open}, Bullet \cite{coumans2015bullet}, PhysX \cite{phsyX}, MuJoco \cite{todorov2012mujoco}, etc.) rely on ``physics engines" comprised of models of real-world processes to predict future states. Unfortunately, due to the assumptions made in constructing the computationally efficient and compact models used in these simulators, predictions made may deviate from real world observations. In particular, \cite{chavan2017,bauza2017probabilistic,Fazeli2017contact} have shown that common contact models in robotics suffer from significant errors when attempting to reproduce real-world observations. Further, the majority of these simulators are deterministic and do not explicitly reason over uncertainty. 

\begin{figure}
    \centering
    \includegraphics[width=\columnwidth]{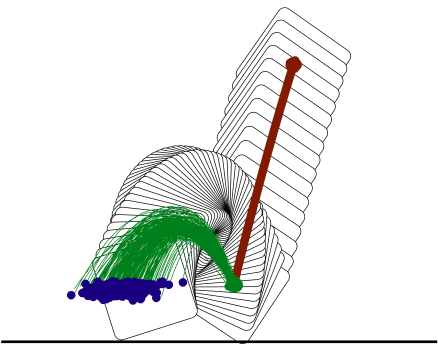}
    \caption{Empirically measured trajectory of the first two contacts of a dice roll, overlayed with samples from the posterior probability distribution from the stochastic simulation. Samples from the initial state of the dice (\textcolor{red}{red points}) are propagated to contact. Contact introduces uncertainty, as shown by the spread in the next set of trajectory samples (\textcolor{green}{green points}). The next contact further increases this uncertainty (\textcolor{blue}{blue points}) and the process continues until the dice comes to rest.}
    \label{fig:dice-uncertainty}
\end{figure}

One approach to dealing with sensitivity to initial conditions, difficulty in computationally efficiently modeling dry friction, and the hybrid nature of contact is to produce simulators that exploit experimental data beyond the scope of parameter optimization (system identification) and instead augment their physical models with the data recorded during experimentation for more accurate predictions. Further, due to the multi-model outcomes, simulators will benefit greatly from tracking beliefs over the set of possible outcomes as opposed to single point predictions. The accuracy and belief propagation properties of these data-augmented models can be used to reduce the burden on controls and planning algorithms. 

In this paper, we build on the data-augmented contact models proposed in \cite{fazeli2017learning} and contribute:
\begin{enumerate}
    \item self-supervised residual learning approach for stochastic long-horizon prediction,
    \item implementation of the approach in pyBullet \cite{coumans2015bullet},
    \item empirical evaluation and bench-marking on a planar impact task.
\end{enumerate}

Here, we use the term data-augmented model to refer to an analytical model cascaded with a ``residual learner" (a data-driven model that makes corrections to the output of the analytical model). The residual learner makes corrections at the force/impulse level and allows for more expressive models with larger predictive ranges than purely analytical ones. The notion of uncertainty is captured by the posterior probability distributions from the residual learner, quantifying the confidence in predictions. 

We first present the simulation framework with an illustrative example. We then discuss the step-by-step implementation of the approach. We conclude the paper with results on an empirical data-set and a discussion of related work. The code and data for this paper is hosted at \texttt{https://github.mit.edu/mCube/residual} \texttt{-pybullet-public}.

\section{Stochastic Simulation Framework}\label{sec:stoch-sim}

The stochastic simulation has two methods; one to propagate states and the other to propagate beliefs. In this particular implementation, we assume that the main source of uncertainty is due to contact interactions, and contact-free motion happens deterministically. As a consequence of this assumption, a ``dynamic step" (state propagation step) in our simulation is a transition from the end of one contact event to the end of the next contact event.

At a contact event we invoke a data-augmented contact model (details in sec.~\ref{sec:data-aug}) which resolves contact while injecting some uncertainty into the current belief over states. We propagate the belief over the states through the dynamic step using a sample efficient hybrid particle/Bayes filter with checks to maintain realizability of predictions (details in sec.~\ref{sec:uncertainty-prop}). 

The main purpose of the stochastic simulation is to make long horizon predictions with uncertainty using data-augmented contact models that make single-step predictions. In the next subsection we provide an illustrative example to walk through the two major steps of the simulation.

This framework is analog to the deterministic simulation framework in the sense that contact-free states are integrated until contact, then contact is resolved and state integration is continued. The distinction here is in the more accurate contact resolution using data-augmented contact models and a propagated belief over states which is directly impacted by the contact model confidence. 

In this study, the uncertainty introduced at contact goes beyond defining probability distributions over the parameters of contact models and instead reasons at the force/impulse level which allows a larger predictive range while capturing some of the unmodeled dynamics of the interaction. This is important since, as \cite{fazeli2017learning} have shown, frictional contact interaction outcomes exist that cannot be predicted by analytical models for any choice of their parameters.

\subsection{Illustrative Example:}

For the empirical evaluation of our simulation paradigm we predict the outcome of planar dice rolls. We provide the initial noisy position and velocity of the dice and simulate forward. Fig.~\ref{fig:dice-uncertainty} shows the first two contacts of an empirical dice roll, and overlayed are samples from the posterior distribution predicted by the stochastic simulation.  

The contact-free motion is governed by gravity and is deterministic, the first dynamic step is then from the initial pose to the end of the first contact, and the second dynamic step is from the end of the first contact to the end of the second contact. At contact the data-augmented contact model injects a measure of uncertainty into the belief over the states of the object that are then propagated along the trajectory. The propagated beliefs become more uncertain as contact occurs. In the following sections we discuss the details of the data-augmented contact models and the propagation mechanisms.

\section{Data-Augmented Contact Models} \label{sec:data-aug}

In this section we first briefly review the key elements of the data-augmented contact models used in this study. These models are based on \cite{fazeli2017learning} and are constructed by cascading an analytic contact model with a ``residual learner". Next, we discuss the transformation necessary to implement these models in pyBullet and similar contemporary physics simulators.

\subsection{Analytical Contact Model:} \label{sec:analytical-model}

Rigid-body contact models resolve contact by selecting a set of compatible contact impulses and post contact states, given pre-contact states and contact parameters for a set of rigid-bodies.

Following \cite{fazeli2017learning}, we denote the state vector as $\myvec{s}=(\myvec{q},\myvec{v})$, were $\myvec{q}$ and $\myvec{v}$ represent configuration and velocity, respectively. We can write the contact model as the following map:
\begin{align}
    \myvec{p}_{m} = \bm{m}(\myvec{s}^{pre}, \mu, \epsilon, \mymat{M}, \myvec{f}_{ext})
\end{align}
where $\myvec{p}_{m}$ is the predicted impulse from the model, $\mymat{M}$ is the inertia matrix of the object, and $\myvec{f}_{ext}$ is the sum of applied external forces. The model selects $\myvec{p}_{m}$ such that the post-contact states satisfy the constraints of rigid-body contact models (decrease in energy, no penetration, linear impulse, principle of maximum dissipation \cite{Stewart:2000}). 

This mapping is parameterized by $(\mu, \epsilon)$, material parameters referred to as the coefficients of friction and restitution respectively. These parameters regulate the magnitudes of the tangential and normal impulses imparted to the object and are coarse lumped-parameter approximations to the complex interactions between the surfaces of objects.

Once the impulse is selected by the model, the subsequent states of the object are computed using the dynamic equations of motion for the instant before and after impact:
\begin{align}
    \myvec{v}^{post} = \myvec{v}^{pre} + \mymat{M}^{-1}(\myvec{f}_{ext}dt + \mymat{J}^T \myvec{p}_{m})
\end{align}
where $\mymat{J}$ is the contact Jacobian. Rigid-body contact models are computationally efficient approximations to a complex phenomenon that occurs over a small but finite time interval where bodies undergo some deformation. 
The models use two parameters to coarsely approximate the interaction and predict a single linear impulse applied at the point contact, as a proxy to a continuum of forces over the actual finite contact time.

Recently, \cite{fazeli2017ISRR} empirically showed that these contact parameters exhibit large variances for a given task. Further, this study suggested that the predictive range of these models was only large enough to explain approximately 50\% of realized planar impacts. Given these limitations, the purpose of the residual learner is to fill the gap between the predictions made by the analytic model and real-world observations.


\subsection{Correcting Point Contact Models}\label{sec:res}

We use a residual learner to correct the impulses predicted by the contact models before passing them onto the dynamic equations of motion, i.e.:
\begin{align} \label{eq:newton}
    \myvec{v}^{post} & = \myvec{v}^{pre} + \mymat{M}^{-1}(\myvec{f}_{ext}dt + \mymat{J}^T \myvec{p}_{cor})
    \\
    \myvec{p}_{opt} & = \myvec{p}_{m} + \myvec{p}_{res} 
\end{align}
where the optimal and residual impulses are denoted by $\myvec{p}_{opt}$ and $\myvec{p}_{res}$, respectively. In our prior work \cite{fazeli2017learning}, we first computed the optimal impulse that best explains a given contact event, then used the error in the estimated impulse from the simulator as the residual to learn. To compute the optimal impulse we pre-multiplying the equations of motion with the contact Jacobian:
\begin{align*}
    \mymat{J}(\myvec{v}^{post} & = \myvec{v}^{pre} + \mymat{M}^{-1}(\myvec{f}_{ext}dt + \mymat{J}^T \myvec{p})) \\
    \myvec{v}_c^{post} & = \myvec{v}_c^{pre} + \mymat{J}\mymat{M}^{-1}\myvec{f}_{ext}dt + (\mymat{J}\mymat{M}^{-1}\mymat{J}^T) \myvec{p} \\
    \Delta \myvec{v}_c & = \mymat{J}\mymat{M}^{-1}\myvec{f}_{ext}dt + \mymat{M}^{-1}_c \myvec{p}
\end{align*}
where $\mymat{M}_c = (\mymat{J}\mymat{M}^{-1}\mymat{J}^T)^{-1}$ denotes the projection of the inertia matrix to the contact frame. The optimal impulse is the solution to:
\begin{argmini*}|s|
{\myvec{p}}{|| \Delta \myvec{v}_c - \mymat{J}\mymat{M}^{-1}\myvec{f}_{ext}dt - \mymat{M}^{-1}_c \myvec{p}||_2}
{}{\myvec{p}_{opt}=}
\addConstraint{\myvec{p}_{n} \geq 0, \; |\myvec{p}_{t}| \leq \mu \myvec{p}_{n}}
\end{argmini*}
where the first constraint prevents penetration and the second enforces Coulomb friction. For multi-point contacts, we can also use the method of impulses as outlined by \cite{Chatterjee:1999}. To learn the residual, we used a Gaussian Process (GP) to represent the model and solved the regression:
\begin{align*}
    \myvec{p}_{opt} = \myvec{p}_{m} + \myvec{p}_{res} \sim \bm{\text{GP}}(\bm{p}_{m}, \; \mymat{K}_{xx}(\myvec{\theta}))
\end{align*}
where $\myvec{\theta}$ denotes the parameters of the GP. Intuitively, the residual learner minimizes the error in post-contact velocity of the center of mass. Here, the contact model parameters are fixed; however, the same optimization holds for joint inference of the GP and model parameters. This formulation requires isolating contact events and estimating the pre- and post-contact states. Further, the corrections are local to the event and agnostic to long-horizon trajectory predictions. To address these limitations, we propose a self-supervised approach to residual model learning that considers the entire trajectory. 

\subsection{Self-supervised Residual Learner}

The objective of the residual learner is to minimize the discrepancy between observed trajectories of an object undergoing contact events and estimated trajectories from a simulator by making corrections to its contact impulses. We denote the $i^{th}$ observed and estimated trajectories as $\myvec{\tau}_i^o = \{\myvec{s}^o_{1...n}\}$ and $\myvec{\tau}_i^e$, and the residual learner policy at time $t$ as $\myvec{p}_{cor, t} = \myvec{\pi}(\myvec{x}_t| \myvec{\theta})$ where $\myvec{\theta}$ are the policy parameters and $\myvec{x}_t$ are the features (details in next subsection). The objective function we use is:
\begin{argmini}|s|
{\myvec{\theta}}{\mathbb{E} \left[\sum_{i=1}^N || \myvec{\tau}_i^o - \myvec{\tau}_i^e ||_2 \right]}
{}{\myvec{\pi}^*=} \label{eq:loss}
\end{argmini}
where the expectation accounts for policy stochasticity and we use the procedure outlined in Alg.~\ref{alg:1} to generate $\myvec{\tau}_i^e$.
\begin{algorithm}
 \KwData{$\myvec{\tau}_i^o$, $\myvec{\theta}$}
 \KwResult{$\myvec{\tau}_i^e$}
 initialization: $\myvec{s}_1^e$ = $\myvec{s}_1^o$, $\myvec{\tau}_i^e = [\myvec{s}_1^e]$ \;
 \For{$t$ in range(1, n - 1):}{
  \eIf{collision($\myvec{s}_t^e$)}{
   $\myvec{a}$ $\leftarrow$ getContactLocation($\myvec{s}_t^e$)\;
   $\mymat{J}$ $\leftarrow$ getContactJacobian($\myvec{a}$)\;
   $\myvec{x}$ $\leftarrow$ getFeatures($\mymat{J}$, $\myvec{s}_t^e$)\;
   $\myvec{p}_{res}$ $\leftarrow$  $\text{sample}(\myvec{\pi}_t(\myvec{x}|\myvec{\theta})$) \;
   $\myvec{p}_{m}$ $\leftarrow$  $\text{getSimImpulse}(\myvec{s}_t^e)$) \;
   }{
   $\myvec{p}_{res}, \; \myvec{p}_{m}$ $\leftarrow 0, \; 0$ \;
  }
  $\myvec{s}^e_{t+1} = \text{simulateStep}(\myvec{s}_t^e, \myvec{f}_{ext} + \mymat{J}^T(\myvec{p}_{res}+\myvec{p}_{m}))$ \;
  $\myvec{\tau}_i^e$.append($\myvec{s}_{t+1}$)
 }
 \caption{Generate $\myvec{\tau}_i^e$}
 \label{alg:1}
\end{algorithm}

\begin{algorithm}
 \KwData{$\myvec{\tau}_{0...L}^o$}
 \KwResult{$\myvec{\theta}^*$}
 initialization: $ \myvec{\theta} \; \leftarrow \; \text{sample} \; \mathcal{N}(0, \sigma) $ \;
 \For{ep in range(maxIter):}{
    \For{l in range(L)}{
        $\myvec{\tau}_l^e$ $\leftarrow$ Generate($\myvec{\tau}_l^o$, $\myvec{\theta}$) (Alg. 1)\;
    }
    $e$ = loss($\myvec{\tau}^e$, $\myvec{\tau}^o$) (eq.~\ref{eq:loss}) \;
    $\myvec{\theta}$ = update($e$, $\myvec{\theta}$)
 }
 \caption{Train policy $\myvec{\pi}$}
 \label{alg:2}
\end{algorithm}

We use a Density Network \cite{bishop1994mixture} to represent the residual policy. This network models the corrective impulse as a normal distribution with mean and variance conditioned on the pre-contact features. This representation is similar to the GP approach, but does not explicitly require output examples for learning. We train the policy using Alg.~\ref{alg:2} where the update routine uses the gradient-free optimization library ``nevergrad'' \cite{nevergrad}. We found that these methods work better for this problem. This may be because contact events move in time and space and may reduce or increase in number during training. Consequently, the gradient information from the loss is too noisy to learn with effectively.

An alternative to this training routine would be to use reinforcement learning \cite{sutton2018reinforcement}. However; since the policy is called only at contact events, and the policy itself affects when the next event occurs, the system is best described by a semi-Markov Decision Process \cite{sutton1999between} which introduces complications.

\textbf{Features:} We may be tempted to directly use $\myvec{x}_s = (\myvec{s}^{pre}, \mymat{M})$ as the input feature space for the residual learner. However, this feature set is an over-parameterization of the impact space and requires more data than necessary. To illustrate, consider a planar square resting on a horizontal and flat surface. The resultant normal and tangential forces are the same for 90 degree rotations of the square (rotational invariances). Similarly, if the square came into contact with the surface at a vertex, the net tangential frictional force applied is proportional to the tangential velocity of the vertex, whether from angular rotation or tangential velocity of the square (impulse-state invariances). 

Intuitively, the contact map is many-to-one for pre- to post-contact state. However, contact models solve for impulses in the contact frame where the mapping from contact impulse to post-contact state is one-to-one. The contact-impulse space is derived by exploiting invariances in the pre-contact states (see \cite{fazeli2017ISRR} for details). We use the same invariant transforms of pre-contact states to contact space as features for the residual policy. These features are $\myvec{x} = (\mymat{M}_c, \myvec{v}_c)$ where $\mymat{M}_c$ is a symmetric positive semi-definite matrix and can be compactly represented with 3 scalars in the plane and 6 in 3D. $\mymat{M}_c$ captures the relative location of the effective inertia of the bodies with respect to the contact frame. $\myvec{v}_c$ denotes the velocity of the contact point.

In terms of scaling, for a single rigid-body with single-point contact in the plane, the raw state and inertia features yield $\text{Dim}(\myvec{s}) + \text{Dim}(\mymat{M}) = 9$ dimensions. For an articulated rigid-body mechanism with 2 links, we would need 18. The dimension increases significantly with the number of links. For single-point contacts $\myvec{x}$ is invariant to the number of bodies in the articulated rigid-body system making contact, and is always 5 dimensional in plane and 9 dimensional in 3D. This is due to the fact that the effective inertia at contact is always a symmetric positive semi-definite $2\times 2$ matrix and the velocity at contact is always a $2\times 1$ vector \cite{Chatterjee:1998}.

\section{Uncertainty Propagation \& The Stochastic Dynamic Step} \label{sec:uncertainty-prop}

Propagating uncertainty through contact is challenging. Contact usually involves solving a highly constrained and non-linear program with sharp changes in velocity and sensitivity to initial conditions. Further, it has been shown experimentally  \cite{bauza2017probabilistic,Fazeli2017contact} that multi-modality is common due to contact/separation transitions, stick/slip transitions, and proximity to singular contacts. Contact models are inaccurate approximations to these phenomena, therefore, capturing prediction uncertainty is important for decision making. Our residual policy estimates this uncertainty using its covariance. 

To illustrate how uncertainty is propagated from impulses to states, consider a contact event at time $t$, the residual policy predicts the correction:
\begin{align*}
    \myvec{p}_{res} & \sim \myvec{\pi}_t = \mathcal{N}(\bm{\mu}_{res,t}, \bm{\Sigma}_{res,t})
\end{align*}
The net impulse imparted to the object is:
\begin{align*}
    \myvec{p}_{res} + \myvec{p}_m &\sim \mathcal{N}(\bm{\mu}_{res} + \myvec{p}_m, \bm{\Sigma}_{res}) =\mathcal{N}(\bm{\mu}_w, \bm{\Sigma}_{res})
\end{align*}
This distribution is propagated through Newton's equations of motion (eq.~\ref{eq:newton}) resulting in a Gaussian distribution for post-contact velocity:
\begin{align} \label{eq:uncertainty}
    \myvec{v}^{post} &\sim \mathcal{N}(\bm{\mu}_{post}, \; \mymat{J}\mymat{M}^{-T}\Sigma_{res}\mymat{M}^{-1}\mymat{J}^T) \\
    \bm{\mu}_{post} &= \myvec{v}^{pre}+\mymat{M}^{-1}(\bm{f}_{ext}dt +J^T\bm{\mu}_w) \nonumber
\end{align}
Note that the contact model introduces uncertainty in the post contact velocity but does not change the configuration of the object or its distribution. This is direct result of one of the assumptions of rigid-body contact models -- contact is resolved instantaneously and configurations do not change. 

Since eq.~\ref{eq:uncertainty} denotes the posterior distribution over the post contact center of mass velocity, we can directly sample from it. For simulators that do not provide access to contact impulses, we can correct in the velocity space. We simply let the simulator resolve contact, then reset the center of mass velocity to the correction sampled from eq.~\ref{eq:uncertainty}.

To capture the multi-modal distribution of outcomes, we represent the belief as a Gaussian mixture:
\begin{align}
    p(\myvec{s}_{t}) = \sum_{i=1}^K \mathcal{N}(\myvec{s}_{t} | \myvec{\mu}_{t,i}, \bm{\Sigma}_{t,i})
\end{align}
where $K$ is the number of Gaussians, and $\bm{\mu}_i$ and $\bm{\Sigma}_i$ are their mean and covariance, respectively. We then propagate the belief by evaluating the integral:
\begin{align}\label{eq:bayes}
    p(\myvec{s}_t) = \int p(\myvec{s}_t|\myvec{s}_{t-1}) p(\myvec{s}_{t-1}) d\myvec{s}_{t-1}
\end{align}

The conjugacy between the normal distributions predicted by the residual learner and the Gaussian mixture allows us to efficiently propagate belief using a hybrid particle Bayes filter. Starting from the Gaussian mixture, at the start of every dynamic step we take $m$ samples, then propagate the samples until contact, then invoke the data-augmented contact model, and finally update the state belief distribution using the weighted Gaussian mixture, then repeat for the next dynamic step.

This approach is a modified version of the belief propagation step of the GP-SUM algorithm ~\cite{bauza2017probabilistic}. The modification we make is to guarantee feasibility of the predictions of the data-augmented contact model. The residual learner does not guarantee that the distribution satisfies physical constraints such as non-penetration and energy balance. To maintain feasibility, we discard samples that lead to penetration or increase in total energy.

The premise of our proposed simulation scheme is that contact free rigid-body dynamics are deterministic and that the main source of uncertainty is contact.
As such, contact events are pivotal in propagating and generating beliefs. A ``dynamic stochastic step" in this simulation framework is then a transition from post-contact states and beliefs of a system to the next post-contact states and beliefs. This macro-step is a measure between two contact events, and is different from the deterministic simulator time-step. In a deterministic simulator, the time-step drives the simulation forward, and is also used here during the deterministic phases of motion.

\section{Experiments}

\begin{figure}[t]
    \centering
    \includegraphics[width=0.95\columnwidth]{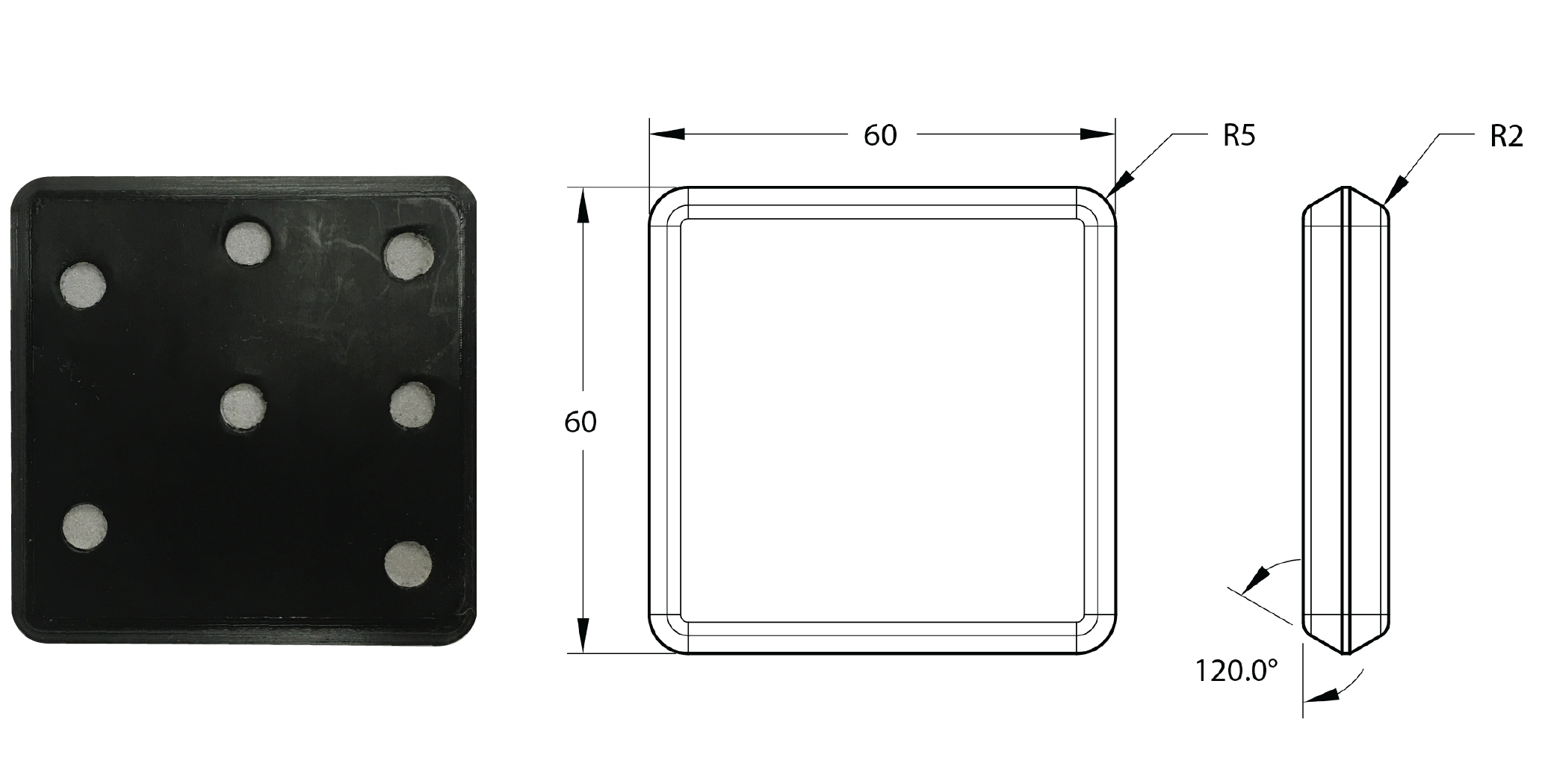}
    \caption{The experimental dice made of PLA and weighs 0.049 kilograms. The dimensions are provided in millimeters. The small tapers are designed to ensure the contact impulses are passed through the center of mass. The dice is covered in a layer of Teflon to minimize lateral friction between the object and the dropping arena. The asymmetric reflective pattern is used by the motion tracking system to collect pose data at 250 hertz.}
    \label{fig:dice-exp}
\end{figure}

To validate the simulation approach, we use empirically collected planar impact data. The dice, Fig.~\ref{fig:dice-exp}, is imparted with random initial conditions within a planar dropping arena replicated from \cite{fazeli2017learning}. The data-set is composed of 500 drop trajectories collected at 250 hertz. Each trajectory is composed of configurations $\myvec{\tau}_i^o=(\myvec{x}, \myvec{z}, \myvec{\theta})_{1...n}$, corresponding to motions in the plane.

\begin{figure}
    \centering
    \includegraphics[trim=0 55pt 0 0, clip, width=\columnwidth]{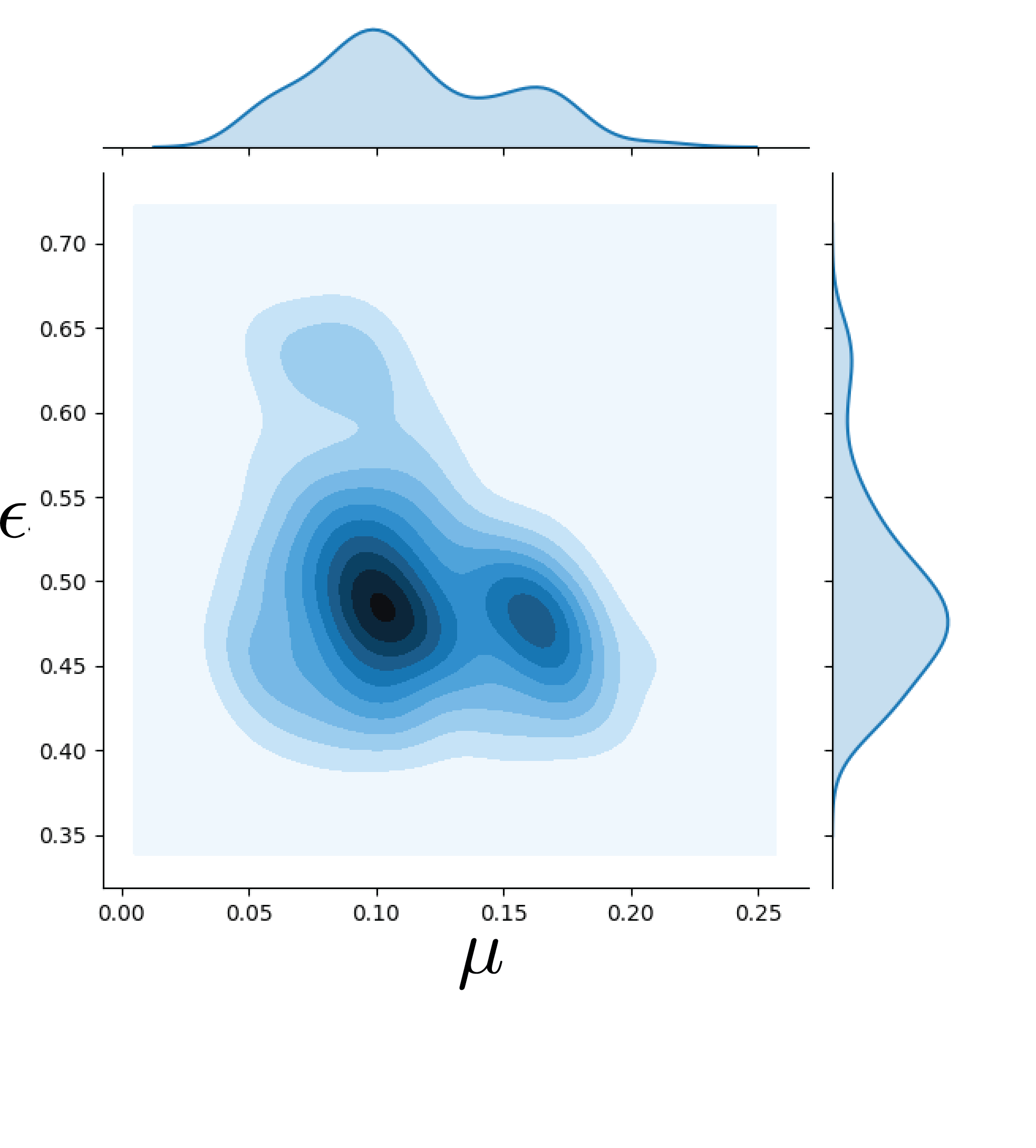}
    \caption{Approximated joint distribution of the contact parameters. The distribution is estimated using the Metropolis-Hastings algorithm. We note the relatively flat distribution for the coefficient of friction, and an almost uni-modal distribution for the coefficient of restitution, similar to the findings of \cite{fazeli2017ISRR} in a similar planar impact task.}
    \label{fig:param-inf}
\end{figure}

\subsection{Simulation Environment and System Identification}

The simulation environment is constructed in pyBullet to replicate the experimental setup as closely as possible. The dice is defined using a Universal Robotic Description Format (URDF). We fix the inertial and geometric properties to match those measured from the experimental dice. The dice has the same six dimensional state vector and the same sample rate as the motion capture system. 

The unknown parameters of the object are its restitution and lateral friction attributes specified by the URDF. To infer the values of these attributes, we perform system identification following the procedure outlined in \cite{fazeli2016parameter}. To this end, we initialize the simulated block with the measured trajectory initial conditions and compute the parameter values that minimize the error.

Fig.~\ref{fig:param-inf} depicts the estimated joint distribution of the lateral friction and restitution parameters. The joint distribution is constructed using the Metropolis-Hastings algorithm with proposal distribution $\mathcal{N}(\myvec{r},\; 0.05 \mathcal{I})$ where $\mathcal{I}$ and $\myvec{r}$ denote the identity matrix and current parameter estimates, respectively. We compute the log-likelihood error using the mean square error loss over the trajectory. To account for disparity in magnitude difference between angular and linear components, we define the trajectory in configuration space as $\myvec{q} = (x, z, r_g \theta)$ where $r_g$ denotes the radius of gyration. 

The algorithm computes 3000 samples with a burn-in period of 1000. For the purposes of simulation, we take the median values of parameters from the joint distribution as point values. These values are $\mu=0.11$ and $\epsilon=0.48$ for the lateral friction and restitution, respectively.

The distribution of parameters matches the findings reported in \cite{fazeli2017ISRR}, where a similar experiment was conducted with an ellipse shape. The median values of the parameters for the ellipse shape were approximately $\mu=0.10$ and $\epsilon=0.52$, in close agreement with the square. This may be explained by the fact that both shapes are made of the same material and are about the same size.

\subsection{Residual Learning}

Once optimized, we use the simulator as the analytical model for the residual learner. To build the training set of the residual learner, we query the simulator for given empirical impacts and compute the correction in the contact frame for each. We record the corrections and features and train the model.

To evaluate the residual learner performance, we initialize the dice in the simulator using test set initial conditions and evaluate the mean square error in predictive performance over the entire length of the trajectory with the same loss term used for system identification. The training set is randomly chosen from the first 400 impact events, and the test set is the remaining 100 impact events. 

\begin{figure}
    \centering
    \includegraphics[width=\columnwidth]{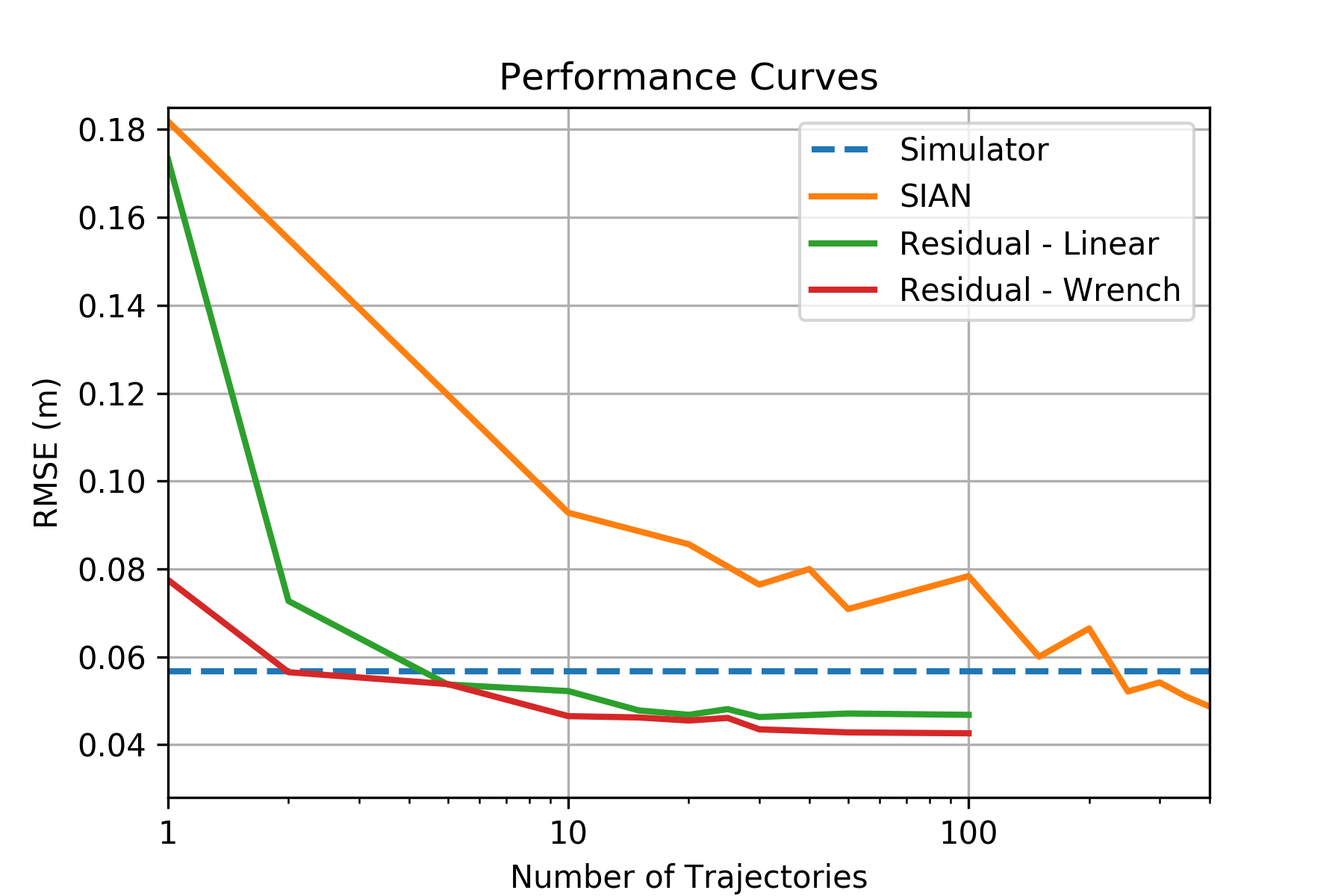}
    \caption{Performance curves for \textcolor{blue}{simulator}, \textcolor{green}{residual linear impulse}, \textcolor{red}{residual wrench impulse} (see \cite{fazeli2017learning}), and \textcolor{orange}{SIAN}. The vertical axis denotes the root mean square error in meters (along the entire trajectory) and the horizontal axis denotes the number of trajectories used to train each model. The point contact residual learner models is significantly more sample-efficient when compared to SIAN, a state-of-the-art general purpose residual learner.}
    \label{fig:residual-exp}
\end{figure}

Fig.~\ref{fig:residual-exp} shows model predictive performance on unseen data as a function of the number of training samples. The models are residual learners for linear impulses and wrenches (see \cite{fazeli2017learning} for details) together with a benchmark algorithm, SIAN \cite{ajay2019combining}. SIAN is a general purpose recurrent neural network formulation for residual learning that has seen success in challenging empirical planar manipulation tasks.

The difference in sample-complexity between the two models is due to their relative size and method of operation. The residual point-contact learners only applies a correction at contact while SIAN attempts to correct state errors at every time-step, including the parabolic free-flight phases -- SIAN is agnostic to contact and requires significantly many more parameters. This result suggests that significant sample-efficiency can be acquire by leveraging physical structure (such as contact) for classes of problems.

Fig.~\ref{fig:residual-exp} also shows the performance of the simulator without augmentation. The point contact residual learners outperforms the simulator between 5-10 training trajectories and plateau at $\approx 25-30\%$ improvement in RMSE over the length of the trajectory. SIAN uses almost the entire available training set to reach simulator performance. 
\section{Related Work}

Learning corrective models has seen renewed interest in robotics. Several works have addressed variants of physical models followed by corrective models \cite{ajay2018augmenting, ajay2019combining, fazeli2017learning, jiang2018data}. We compare the performance of the approach proposed here to the first two works by extending the third work. In \cite{jiang2018data} the authors also propose an interesting residual method. It is unclear how well the approach works for long-horizon predictions but is a promising direction.

In related work, \cite{byravan2017se3} showed how to design a neural network to predict rigid-body motions in a planar pushing scenario. In this study, the neural network differentiates between object and table as a robot pushes the object. The neural network makes predictions by explicitly predicting $SE(3)$ transformations. This technique allows a compact representation of the states of the object and prevents ``blur" of the edges of the object.
In a similar planar pushing task, \cite{kloss2017} propose an analytic pushing model ``informed" by a neural network to out-perform both purely data-driven and purely analytic models in single-step predictions of the outcome of a push. The model is learned from a large pushing data-set~\cite{Yu2016}. 

In \cite{zhou2016convex}, the authors provide a data-efficient approach to model the frictional interaction between an object and a support surface
by directly approximating the mapping between frictional wrench and slipping twist.
In later work~\cite{zhou2017fast} they extend the model to simulate parametric variability in planar pushing and grasping.

One interesting approaches to correcting models is shown by \cite{allevato2019tunenet}, where corrections are made to model parameters. It is unclear how well this approach generalizes to tasks where the simulator cannot reproduce physical interactions for any choice of parameters. In \cite{silver2018residual}, the authors learn a residual on a policy rather than a physics model, providing a novel approach to bootstrap learning in policies.

\section{Discussion - No Free Lunch}

In this paper, we demonstrated one approach to more accurate long-horizon simulation and uncertainty propagation through contact using residual learner models specialized for point contacts. These models exploit physics knowledge (contact modeling) to significantly improve sample-efficiency when compared to general purpose techniques.

The improved efficiency comes at a cost. A significant engineering effort is required to process the empirical trajectories, detect contacts, and compute impulse corrections. This is in contrast to SIAN, where the only requirement is to provide a simulated and empirical trajectory. 

An artifact of choosing the impulse from a smooth and continuous distribution is that the model cannot predict certain outcomes such as sliding to sticking or perfect sticking interaction as these modes have a zero measure under these distributions. Effectively, the residual learner blurs the line between the two distinct modes of sticking and sliding. One potential remedy to this issue is the use of a supervisory algorithm that decides when a predicted impulse is ``close enough" to resulting in sticking and projecting the impulse onto the line/plane of sticking impulses.

Another technical issue comes from the choice of belief representation and belief propagation. In this study we use a Gaussian Process model as residual learner and represented the belief with a mixture of Gaussians. Due to the infinite support of Gaussian distributions, a portion of the belief will always be infeasible, e.g. regions in penetration. In this study we found empirically that this portion of the belief was very small, and it was sufficient to filter spurious samples to keep fully feasible beliefs.

Here, we did not consider multi-point contact events. These scenarios seem challenging since multi-contact events are often indeterminate, i.e. many impulses exist that explain the interaction. In this case, what is the correct residual to learn? Further, while the model does account for some deformation, it is not clear how well it will perform on deformable bodies.

The results of this study suggest that, when available, specialization of the residual learner is extremely beneficial. However; we do not yet have a suite of residual learners for broad classes of tasks involving frictional interaction. 

\addtolength{\textheight}{-12cm}   






\bibliographystyle{IEEEtran}
\bibliography{IEEEabrv,example_paper}

\end{document}